\documentclass[11pt]{article}

\usepackage{jhu_paper}
\date{}
\usepackage{array}
\usepackage{colortbl}
\usepackage{tabularx}
\usepackage{multirow}
\usepackage{subcaption}
\usepackage{enumitem}
\usepackage{wrapfig}
\usepackage[numbers,sort&compress]{natbib}
\newcommand{\method}{MotionForesight}
\newcommand{\trackcraft}{TrackCraft3R}
\newcommand{\wan}{Wan2.1}
\newcommand{\Kctx}{K}

\newcommand{\DefaultK}{7}
\newcommand{\DefaultN}{15}
\newcommand{\DefaultT}{22}
\newcommand{\TrainVideos}{40K}
\newcommand{\TrainableParams}{12.19M}
\newcommand{\ModelRes}{$320\times576$}

\definecolor{BestCellBlue}{RGB}{232,244,255}
\definecolor{SecondCellBlue}{RGB}{244,249,255}
\newcommand{\bestcell}[1]{\cellcolor{BestCellBlue}\textbf{#1}}
\newcommand{\secondcell}[1]{\cellcolor{SecondCellBlue}#1}

\title{MotionForesight}
\jhusubtitle{Re-purposing Video Models for Future 3D Scene-Flow Prediction}
\author{Homanga Bharadhwaj$^*$ \quad Yash Jangir$^*$\\[0.2em]
\footnotesize{($^*$ authors contributed equally)}}
\jhuinstitution{Department of Computer Science, Johns Hopkins University}
\jhucontact{\texttt{hbharad2@jhu.edu} \quad \texttt{yjangir@jhu.edu}}
\jhuvenue{Brains, Bots, and Behavior Lab}
\jhurunningtitle{MotionForesight}

\begin{document}
\maketitle

\begin{figure}[h!]
    \centering
    \vspace*{-0.6cm}
\includegraphics[width=0.93\linewidth]{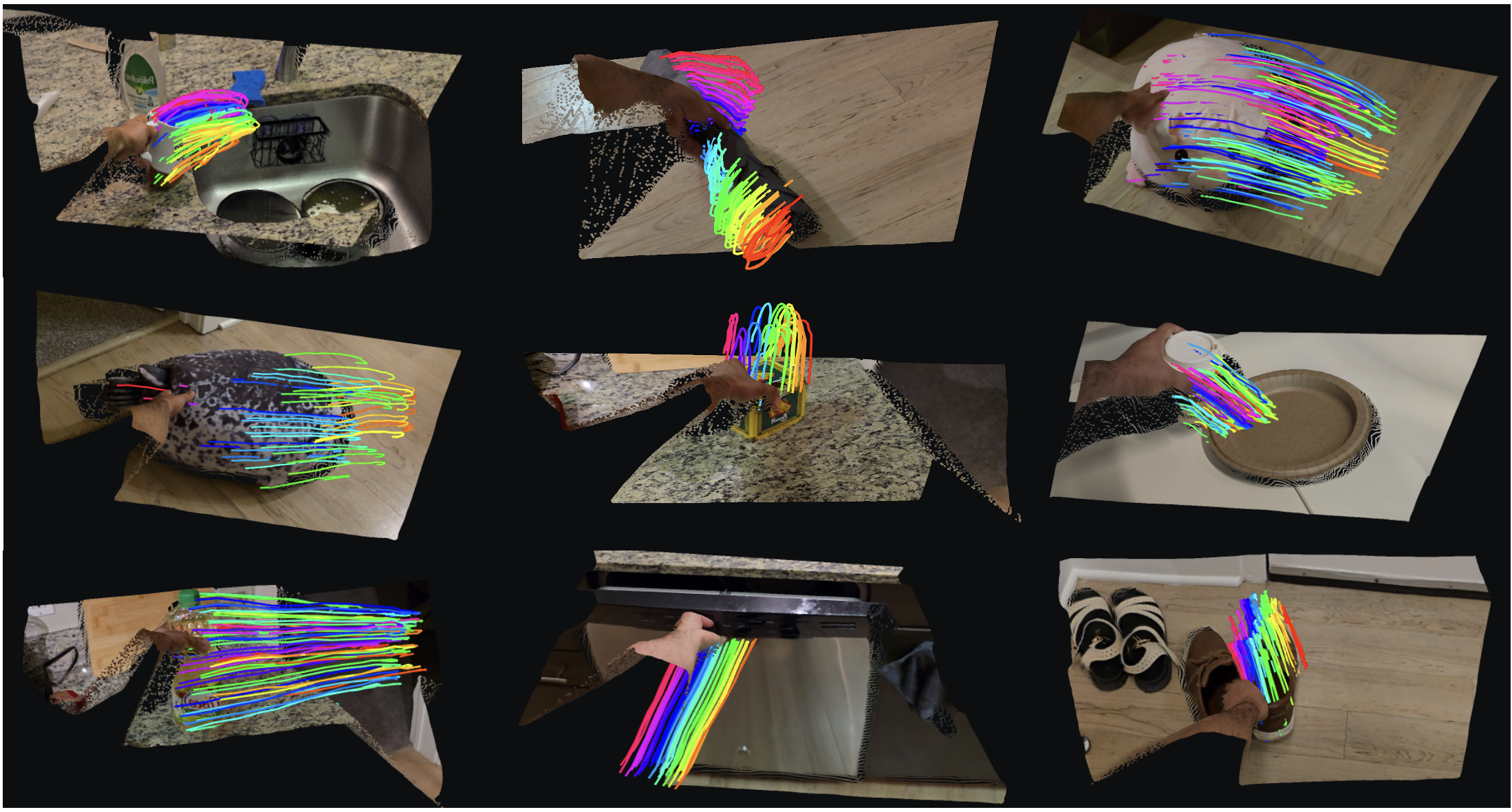}
    \caption{MotionForesight forecasts plausible motion in diverse everyday manipulation scenarios}
    \label{fig:placeholder}
    \vspace{-0.4cm}
\end{figure}
\begin{abstract}
Humans can infer how objects are likely to move from passive observation: a cup may be lifted, a drawer may slide, and a lid may rotate shut. Such predictions expose the physical consequences of interaction needed to act in the real world. We study how to learn this anticipation from ordinary monocular videos of human-object interaction. Given a short observed video context, MotionForesight predicts future 3D trajectories for points on the manipulated object. This casts interaction prediction as object-centered 3D motion forecasting without any assumptions on the object properties. Our key insight is that video prediction models already encode rich priors about how objects move during human interactions. We redirect these priors from pixel prediction toward future 3D scene flow. We start from a dense 3D tracker built on a pretrained video model, generate pseudo-ground-truth tracks from complete clips, and train the forecaster using only the observed frames. We replace future RGB and geometry with learned mask latents and train a lightweight adapter to turn the retrospective tracking representation into a forward predictor, while freezing the large video and tracking components. Using just \TrainVideos{} human videos and no auxiliary inputs such as language, MotionForesight generalizes across diverse out-of-distribution objects, environments, viewpoints, and interactions. It also outperforms substantially larger models that use over a million training videos. These results show that we can efficiently re-purpose video priors into explicit geometric forecasts for embodied intelligence. \href{https://motionforesight.github.io/}{motionforesight.github.io}
\end{abstract}

\section{Introduction}
\begin{quote}
\vspace{0.2cm}
\centering
\emph{``One of the most fundamental properties of thought is its power of predicting events.''}\\
---Kenneth Craik, \emph{The Nature of Explanation}
\end{quote}
A central component of embodied intelligence is not only recognizing what has happened, but \textit{anticipating what will happen next}. When people watch a hand approach a mug, a knife press into a fruit, or a cabinet door begin to move, they often infer the likely future configuration of the object before the motion is complete. This ability goes beyond visual prediction by reflecting an understanding of object affordances, contact, constraints, and goals. Studies in cognitive science have long emphasized that perception is tied to action possibilities \citep{gibson1979ecological}, that motion supports structured scene interpretation \citep{ullman1979interpretation}, and that humans often reason by mentally simulating physical futures \citep{battaglia2013simulation}. The classic rational-imitation result of \citet{gergely2002rational} further suggests that infants do not simply do as we do,  but interpret observed actions in light of goals and constraints. \textbf{Modeling interaction effects} is therefore crucial: what matters for observational learning is how the object will move given a specific context, not the exact hand motion that produced it.

In this paper, we tackle the problem of \textit{forecasting motion} in the form of \textit{future 3D scene flow} for manipulated objects from casual monocular human videos. 3D scene flow provides a general representation for inferring motion in scenes: instead of committing to a category-specific state such as a rigid pose, it describes how points in the 3D scene move over time. Our setting differs from related directions along two main axes: the information used to specify the future and the representation used to describe motion. Some methods rely on language or action grounding to predict either sparse 3D point trajectories~\citep{molmomotion} or rigid 6-DoF object-manipulation trajectories~\citep{language6dof}. Others predict future rigid 6-DoF object poses directly from visual observations~\citep{objectforesight}. Although pose representations are compact for rigid objects, they cannot naturally describe articulated parts, deformable surfaces, or local nonrigid motion. Some prior works predict future 2D point tracks for robot manipulation~\citep{track2act,atm,motiontracks}, whereas our target is explicit future metric 3D motion rather than image-plane tracks in a downstream control setting. We focus on dense, reference-anchored 3D scene flow because it provides a general interface for interaction dynamics: it is metric rather than image-plane, object-centered rather than embodiment-specific, and not restricted to a single rigid pose parameterization. In principle, the same representation can describe rigid objects, articulated parts, deformable surfaces, and local object motion. \textit{Learning this prediction from everyday videos is therefore a compelling problem}: casual videos are noisy and unconstrained, but they contain broad evidence about how objects move when people use them.

\begin{figure}[t]
    \centering
    \includegraphics[width=0.98\textwidth]{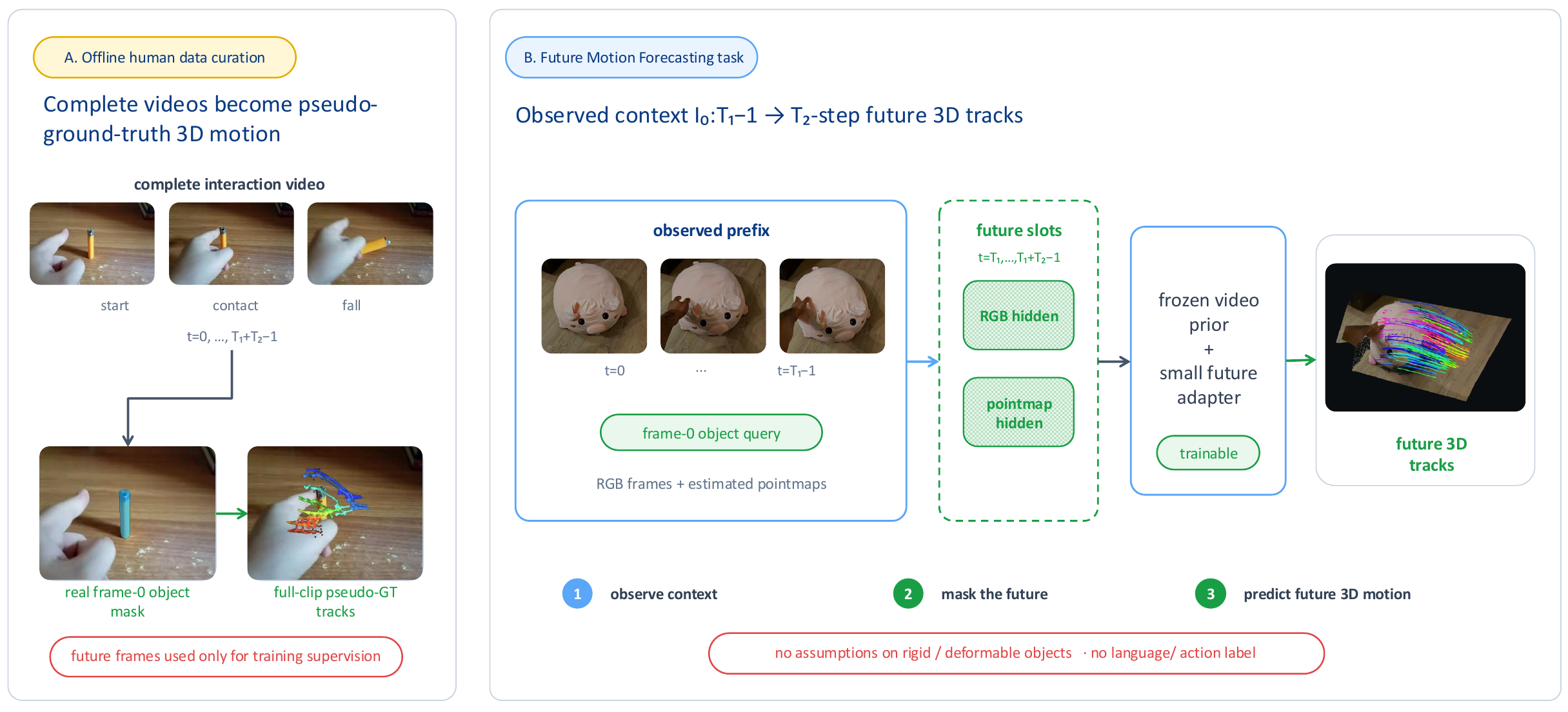}\vspace*{-0.2cm}
    \caption{\textbf{MotionForesight} predicts future reference-anchored 3D tracks from passive human video. Given observed RGB frames and pointmaps obtained through estimated monocular depth, for the first $T_1$ frames, the model forecasts the future motion of the manipulated object for the next $T_2$ steps. The output is an explicit 3D trajectory field over points on the manipulated object and does not require a rigid-body or category-specific motion parameterization.}
    \label{fig:teaser}
\end{figure}
Our key insight is that everyday human videos already describe the motion of common objects we want to predict, while modern video models are likely to encode useful priors about how such motion unfolds. A model trained on large video corpora must represent temporal regularities such as contact, affordance, object persistence, and plausible short-horizon dynamics, even if its native output is a future image or video latent. We ask whether these priors can be redirected toward a more explicit geometric target. Rather than generating future pixels and recovering motion afterward, MotionForesight directly predicts the future 3D trajectory field for object-attached points. The model uses the observed context to infer the likely evolution of the interaction, but its output remains a compact, metric motion representation that can be consumed by downstream planning or embodied reasoning systems. This direct geometric prediction is also computationally attractive for dynamic robot manipulation: it avoids spending inference time rendering future RGB frames and then running a separate tracking or reconstruction pipeline to recover the motion that the robot actually needs. By predicting future 3D motion directly, the model exposes an actionable representation with lower overhead than a generate-then-track alternative.

Concretely, MotionForesight repurposes TrackCraft3R \citep{trackcraft3r}, a feed-forward dense 3D tracker built around a video DiT~\cite{wan2025}. TrackCraft3R is retrospective: it observes all frames, encodes RGB frames and reconstructed pointmaps into geometry latents, repeats a first-frame query latent across time, and predicts reference-anchored tracking pointmaps. We convert this tracker into a forecasting model by hiding the future frames. For timestamps after the observed prefix, the unavailable RGB and pointmap latents are replaced by learned mask latents, while the frame-0 query latent and temporal RoPE interface are preserved. The transformer therefore receives observed 3D context, a reference-frame object query, and future time indices, and it predicts residual track latents that decode into future 3D tracks. We keep the base video model, the original TrackCraft3R adapter, and the VAE encoders/decoders frozen, and train only a fresh low-rank adapter, I/O projections, the prediction head, and the mask latents. Pseudo-ground-truth future tracks are generated by running dense 3D tracking on complete human interaction clips, then training the model with those future observations removed.

\jhukeypoint{MotionForesight uses passive monocular human videos to learn \emph{object-centered future 3D motion}. This formulation abstracts away from the human's motor command and from the appearance of future pixels by focusing on how points on the manipulated object are likely to move.}

\noindent\textbf{In summary, we claim three contributions.} First, we formalize future 3D scene-flow prediction from casual RGB interaction videos as a reference-anchored tracking-pointmap forecasting task. Second, we present a data curation pipeline that converts human videos into pseudo-ground-truth future tracks using dense 3D tracking and object masks, while masking future frames at training time. Third, we introduce a minimal modification to TrackCraft3R~\cite{trackcraft3r} that re-purposes it from a point track extraction model to a future point track prediction model. We show that this surprisingly simple and compute-efficient recipe trained on just 40k RGB human videos achieves compelling motion forecasting results in generic real-world scenes. 

\section{Related Work}

\paragraph{Visual forecasting and world models}
Building models of how the visual world evolves has long been a central problem in computer vision and robotics. Prior work predicts future pixels or latent states for planning, control, and representation learning~\citep{ha2018worldmodels,finn2017visualforesight,villegas2017video,walker2016uncertain}. More recent approaches scale this idea through video diffusion, predictive visual representations, and learned latent-action models~\citep{gen2act,vpp,videopolicy,latentactionworldmodels,vjepa2,dreamer,cswm}. These methods capture useful temporal regularities, but their predictions are typically represented as RGB frames, video features, or action-conditioned latent states without explicit object-level geometry. Our work builds on the temporal priors learned by video models but differs in its output representation. Rather than generating future pixels or abstract features, \method{} predicts the future metric 3D motion of points on the manipulated object.

\paragraph{Inferring interaction cues from human videos}
Large-scale datasets such as Something-Something~\citep{something}, YouCook~\citep{youcook}, EPIC-Kitchens~\citep{epickitchens2018}, EGTEA~\citep{egtea}, and Ego4D~\citep{ego4d} have enabled models to learn human--object interaction priors directly from video. One line of work predicts future action labels, active objects, contact regions, hand trajectories, and interaction hotspots~\citep{liu2020forecastinghoi,hoigaze,contactgrasp,handsprobes,jointhandmotion,where2act,newtonian,interactionhotspots}. Other work uses large-scale activity and contact observations to learn how people interact with objects~\citep{handscontact,cmummac}. More recent methods forecast language- or context-conditioned hand trajectories~\citep{handsonvlm,flowingreasoning}, while robot-learning approaches predict 2D point tracks, masks, or visual tokens as embodiment-agnostic plans~\citep{track2act,atm,mask2act}. These methods extract actionable cues from human videos, but their predictions largely remain semantic, hand-centric, or tied to the image plane. \method{} instead predicts continuous 3D trajectories for points in the scene, directly representing the physical consequence of an interaction without requiring language/action labels.

\paragraph{3D geometric extraction and forecasting}
Another line of work models interaction through explicit geometry. Early approaches estimate 3D hand and object poses from visual observations~\citep{ge2019hand,hasson2019joint,iqbal2018hand,frankmocap,zimmermann2017hand} or recover 6-DoF object pose from RGB and RGB-D inputs~\citep{pvn3d,segdriven6d,ssd6d,posecnn}. Recent systems provide complementary tools for video segmentation and single-image 3D reconstruction: SAM~2~\citep{sam2} propagates masks through video, while TRELLIS~\citep{trellis} and SAM~3D~\citep{sam3d} reconstruct 3D geometry from images. Future geometry prediction has also been studied through LiDAR, point-cloud, and agent-trajectory forecasting in autonomous-driving and mobile-robot settings~\citep{mersch2021pointcloudprediction,vidar,vip3d}. Closer to manipulation, methods predict language-conditioned 6-DoF trajectories~\citep{language6dof}, transfer human trajectories to robot embodiments~\citep{maniptrans}, or forecast object poses, point motion, and scene flow under language, goal, RGB-D, or robot-action conditioning~\citep{objectforesight,pointworld,dexwm}.  In contrast, \method{} predicts dense, reference-anchored dense 3D tracks from monocular RGB videos, allowing the same representation to describe rigid, articulated, and deformable motion. Concurrent with our work, MolmoMotion predicts a closely related output: future metric 3D trajectories, but formulates goal-conditioned forecasting from a short RGB history and a language description of the intended action, and predicts the motion of just 8 points on an object~\citep{molmomotion}. It is trained on MolmoMotion-1M, an action-described, object-grounded corpus constructed from 1.16M diverse videos, whereas \method{} is trained on 40K monocular SSv2 human-interaction videos and receives no language or action input. Thus, the two works share a point-trajectory output representation but study distinct supervision regimes: language-grounded intent from a broad video corpus for sparse points versus passive visual anticipation from observed interaction for dense 3D flow forecasting.

\paragraph{Scene flow and point tracking}
Scene flow and point tracking provide the motion representation underlying our approach. Classical and learned scene-flow methods estimate dense 3D motion between observed frames, while recent trackers recover long-range correspondences through camera motion, occlusion, and nonrigid deformation~\citep{flownet3d,tapir,alltracker,cotracker,delta,spatialtrackerv2,trackcraft3r}. These methods are primarily retrospective: they estimate where points moved after observing the relevant frames. We use this capability both to construct pseudo-ground-truth supervision and to define the output interface of our model. We then turn retrospective tracking into forecasting by hiding future observations and asking a tracking-adapted video backbone to predict where reference-frame object points will move before those frames become available.

\section{MotionForesight}
\label{sec:method}

\method{} aims to predict the future motion of objects being manipulated in a scene from a short video context. Toward this goal, we pose a geometric version of the video-prediction problem. Instead of predicting future images, we predict the future motion of object points that are already visible during the observed interaction. This choice is motivated by embodied settings, where a predictive model must capture changes in scene geometry more precisely than changes in texture or appearance. The physical consequence of an interaction is often better represented by where an object moves in 3D than by how the subsequent RGB frames look.

Concretely, given $T_1$ observed RGB frames, we predict the future 3D tracks of points on the manipulated object for the next $T_2$ time steps. At inference, we observe only the video prefix $I_{0:T_1-1}$. We estimate a pointmap $P_t\in\mathbb{R}^{H\times W\times3}$ for each observed frame and sample query points $\mathcal{Q}$ from the object mask in the reference frame $a=0$. Our goal is to predict the future 3D locations of these reference-frame points:
\begin{equation}
\hat{X}_{t}(q)\in\mathbb{R}^{3},
\qquad q\in\mathcal{Q}, \quad
t=T_1,\ldots,T_1+T_2-1.
\end{equation}

During training, we have access to videos of length $T=T_1+T_2$. We use the complete videos only offline to extract pseudo-ground-truth 3D tracks; the forecasting model itself receives only the first $T_1$ RGB frames and their estimated geometry. We express the observed geometry and future tracks in the last-observed camera frame, $t=T_1-1$, to reduce apparent motion caused by the camera. Our default setting uses $T_1=\DefaultK$, $T_2=\DefaultN$, and $T=\DefaultT$. The model receives no auxiliary inputs like language instructions or action labels.

\subsection{Data curation: from RGB human videos to 3D object trajectories}

We construct pseudo-ground-truth supervision from approximately \TrainVideos{} human-object interaction videos from the Something-Something V2 dataset. Because these videos do not provide metric 3D trajectories or temporally consistent object masks, we process each clip using an offline segmentation, reconstruction, and tracking pipeline.

The manipulated object is not always visible in the first frame, so we select an intermediate anchor frame in which it is sufficiently clear. We use the object name from the dataset annotation as a cue for mask extraction, initialize Segment-Anything~\cite{segmentanything} on the anchor frame, and propagate the resulting mask backward and forward through the clip. Query points are densely sampled within the object mask and retained wherever valid object support is available.

We estimate monocular depth for every frame using DepthAnything3~\cite{da3}, recover camera motion, and combine both estimates to construct temporally aligned pointmaps. We then run \trackcraft{}~\cite{trackcraft3r} on the complete clip to obtain reference-anchored 3D trajectories for the sampled object points. The resulting tracks are transformed into the last-observed camera coordinate frame and stored with their validity masks. The complete clip is used only for offline pseudo-label generation; during training, future RGB frames and pointmaps are never provided to the forecasting model.

\subsection{Observed visual-geometry context}
\label{sec:visual-geometry}
The first modeling step converts the observed video prefix into the latent sequence consumed by the video transformer. For each observed frame $I_t$, we pair the RGB image with its estimated pointmap $P_t$. The RGB and geometry streams are encoded separately and concatenated into a visual-geometry latent,
\begin{equation}
c_t =
\left[
E^{\mathrm{rgb}}(I_t);
E^{\mathrm{pm}}(P_t)
\right],
\qquad t < T_1.
\end{equation}
Here, $E^{\mathrm{rgb}}$ is the RGB VAE encoder and $E^{\mathrm{pm}}$ is the pointmap VAE encoder. Pointmap normalization statistics are computed using only the observed prefix because future geometry is unavailable at inference time. 

The two streams provide complementary evidence to the model. The RGB stream captures appearance, contact, and human-object interaction cues, while the pointmap stream provides metric scene structure and observed 3D motion. Together, they allow the transformer to reason about which object is moving, how it is constrained, and how its visible motion is likely to continue into the future.

\begin{figure}[t]
\centering
\includegraphics[width=0.98\textwidth]{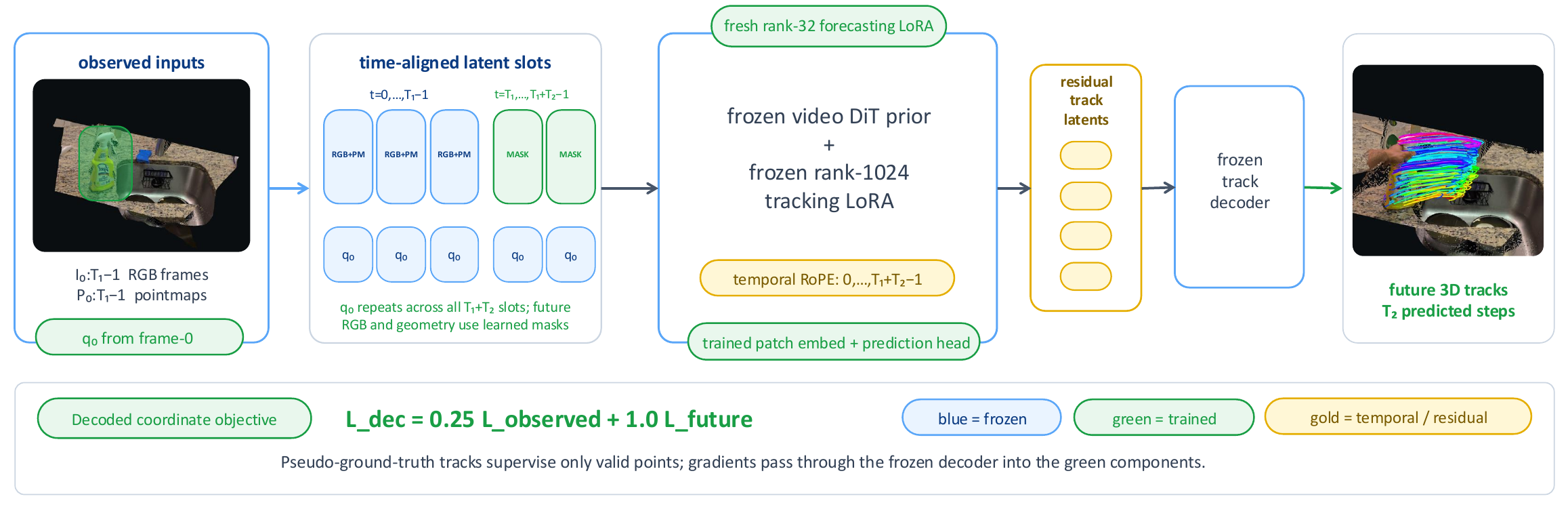}\vspace*{-0.2cm}
\caption{\textbf{Architecture.} We preserve TrackCraft3R's dual-latent construction. Observed frames produce standard RGB and pointmap geometry latents. Future geometry slots are filled by learned mask latents. The frame-0 track/query latent is repeated across all timestamps, and temporal RoPE assigns the target future time. A frozen video DiT with a small fresh LoRA predicts future residual-track latents, which the frozen track decoder converts into future tracking pointmaps.}
\label{fig:method}
\end{figure}

\subsection{A point-track interface for a video prior}

A pretrained video model contains a useful temporal prior, but its native output is not a metric 3D trajectory. We therefore start from \trackcraft{} \citep{trackcraft3r}, a dense 3D tracking model built on top of a video DiT, Wan2.1~\cite{wan2025}. \trackcraft{} provides an already-learned interface between video latents and reference-anchored 3D point tracks.

For a fully observed tracking clip, \trackcraft{} uses two time-aligned latent streams. The first is the context stream $c_t$ defined above (Section~\ref{sec:visual-geometry}), which varies with each video frame. The second is a reference, or query stream that repeats the reference-frame state at every timestamp:
\begin{equation}
r_t =
\left[
E^{\mathrm{rgb}}(I_a);
E^{\mathrm{pm}}(P_a)
\right],
\qquad
a=0,\qquad
t=0,\ldots,T-1.
\end{equation}
Repeating the query stream turns every temporal output slot into the same geometric question: \textit{where is each reference-frame point at time $t$?} The video transformer receives the context and query latents together with temporal RoPE and predicts a residual-track latent $\hat{z}^{\Delta}_t$. A frozen track decoder maps this latent to a 3D residual, which is added to the reference point:
\begin{equation}
\hat{X}_t(q)
=
X_0(q)
+
D^{\mathrm{track}}(\hat{z}^{\Delta}_t)(q).
\end{equation}
Each output therefore has a direct geometric interpretation in terms of the predicted 3D position of a reference-frame object point. The pointmaps $P_t$ are obtained by unprojecting per-frame depth and transforming the resulting 3D points into the reference-camera coordinate system. Thus, all geometry inputs share a common frame—an essential property that enables the model to produce reference-anchored tracks. \trackcraft{} uses the video DiT as a single-step latent regressor at a fixed regression timestep rather than as an iterative diffusion sampler; \method{} inherits this deterministic single-pass interface when converting tracking into future forecasting.

\subsection{Turning tracking into forecasting}

The original tracking model is retrospective: it observes the complete video and explains where the reference points moved. We convert it into a forecasting model by hiding all future observations. For observed timestamps, the context stream contains the actual visual-geometry latents,
\begin{equation}
\tilde{c}_t =
\left[
E^{\mathrm{rgb}}(I_t);
E^{\mathrm{pm}}(P_t)
\right],
\qquad t < T_1.
\end{equation}
For future timestamps, no RGB image or pointmap is available. We therefore replace both components of the future context stream with learned mask latents,
\begin{equation}
\tilde{c}_t =
\left[
m^{\mathrm{rgb}};
m^{\mathrm{pm}}
\right],
\qquad t \geq T_1.
\end{equation}
The RGB mask latent $m^{\mathrm{rgb}}$ and pointmap mask latent $m^{\mathrm{pm}}$ are shared across future timestamps and broadcast spatially. They do not encode a particular future appearance or geometry. Instead, they indicate that the corresponding temporal slot is unobserved. Temporal Rotary Position Embedding (RoPE)~\cite{rope,rope1} assigns each slot a distinct time index, allowing the transformer to distinguish near- and long-horizon predictions even though the learned unknown-token content is shared.

We repeat the reference query stream across all $T$ timestamps. Consequently, the model receives the observed visual-geometry context, the same reference-object query at every timestamp, and the temporal index of each requested future. It must then infer where the corresponding object points will move. This formulation directly exposes future 3D scene flow without first generating RGB frames and subsequently applying a separate reconstruction or tracking pipeline.

\subsection{Forecasting supervision and objective}

The curated full-clip trajectories provide pseudo-ground-truth targets $X_t(q)$, but the forecasting model receives only the observed prefix, the reference query, and the masked future slots. Future RGB frames and future pointmaps are never included in the model input.

The implementation supports both dense pointmap supervision and sparse query-point supervision. We express both using a common query-point interface. For dense clips, object points are sampled from the propagated reference-frame mask and their targets are read from the dense tracking pointmaps. For sparse clips, we use the native tracked points and their associated validity values. Let $v_{tq}$ indicate whether point $q$ has valid supervision at timestamp $t$.

The default objective is a decoded coordinate-space loss,
\begin{equation}
\mathcal{L}_{\mathrm{dec}} =
\lambda_{\mathrm{obs}}
\frac{
\sum_{t<T_1}\sum_{q} v_{tq}
\left\|
\hat{X}_t(q)-X_t(q)
\right\|_2^2
}{
\sum_{t<T_1}\sum_q v_{tq}+\epsilon
}
+
\lambda_{\mathrm{fut}}
\frac{
\sum_{t\geq T_1}\sum_{q} v_{tq}
\left\|
\hat{X}_t(q)-X_t(q)
\right\|_2^2
}{
\sum_{t\geq T_1}\sum_q v_{tq}+\epsilon
}.
\label{eq:decoded_loss}
\end{equation}
We weight future prediction more heavily than observed-frame reconstruction. The observed term primarily stabilizes the inherited tracking interface, while the future term trains the model to extrapolate object motion beyond the available evidence. The loss back-propagates through the frozen track decoder, but the decoder parameters themselves are not updated.

\begin{figure}[t]
    \centering
    \includegraphics[width=\linewidth]{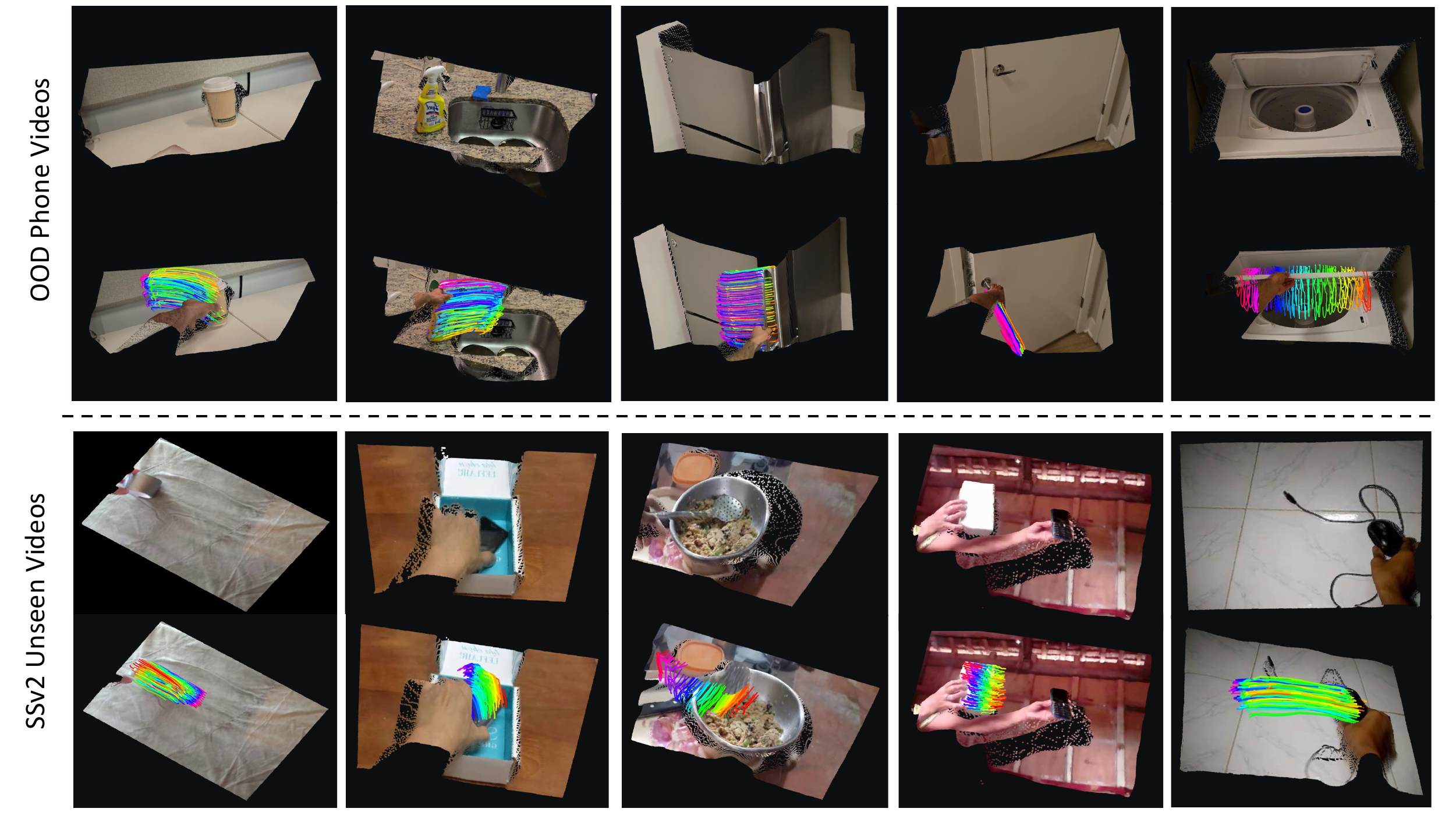}
    \caption{\textbf{Qualitative results for future 3D track predictions.}
    From only the observed video prefix, \method{} predicts plausible object motion including lifting, translation, rotation, constrained sliding, and local nonrigid motion. The visuals show pointmaps of the first and last observed frames, and the future predicted 3D tracks overlaid on the last observed frame. Detailed videos in \href{https://motionforesight.github.io/}{motionforesight.github.io}}
    \label{fig:qualitative}
\end{figure}

\section{Experiments}
\label{sec:experiments}

We organize our experiments around four research questions:

\begin{enumerate}[leftmargin=*, nosep]
    \item \textbf{Re-purposing video priors:} Can a pretrained video backbone be effectively adapted for future 3D track forecasting?
    \item \textbf{Scaling with human video:} How does forecasting improve with more (both quantity and diversity) human-interaction videos?
    \item \textbf{Out-of-distribution generalization:} How well does the learned forecasting recipe transfer beyond the fine-tuning distribution?
    \item \textbf{Role of auxiliary supervision:} How does our approach, trained without language or action labels, compare with methods that use such information?
\end{enumerate}

\subsection{Datasets and baselines}
\label{sec:exp_setup}

We train \method{} on \TrainVideos{} human-object interaction videos from Something-Something V2 (SSv2)~\citep{something}. The model does not receive SSv2 action labels, language instructions, or auxiliary action annotations during either training or inference. Text is used only by the offline preprocessing pipeline, when needed, to identify the manipulated object.

We evaluate on 150 held-out SSv2 videos and 50 independently recorded phone videos. The phone videos in home and office scenes contain previously unseen objects, environments, viewpoints, and capture conditions, providing an out-of-distribution (OOD) evaluation. The complete clip is used only offline to construct pseudo-ground-truth trajectories; every forecasting method observes the same first $T_1$ frames and predicts the following $T_2$ frames.

All methods are evaluated using the same object query points, future timestamps, validity masks, and coordinate frame. Since MolmoMotion allows forecasting only 8 points at once, we do multiple passes through it. Following prior trajectory-forecasting work~\citep{molmomotion,objectforesight,thakkar2026forecasting,flowingreasoning}, we report average displacement error (ADE), final displacement error (FDE), and the percentage of predictions within a fixed distance threshold (PWT). ADE and FDE are reported in centimeters; lower ADE/FDE and higher PWT(@5cm) are better.

\begin{table}[t]
  \centering
  \scriptsize
  \caption{\footnotesize\textbf{Comparison on SSv2 and OOD phone videos.}
  All methods use the same observed interval and prediction horizon. Rows below the horizontal rule receive a
  ground-truth action description and therefore use more test-time information than \method{}. Note that
  MolmoMotion is trained with 1M videos from diverse sources. \method{} is
  trained with 40k RGB videos from SSv2.}
  \label{tab:baselines}
  \resizebox{\linewidth}{!}{%
  \begin{tabular}{llcccccc}
  \toprule
  & & \multicolumn{3}{c}{SSv2 unseen clips (150)}
  & \multicolumn{3}{c}{OOD phone videos (50)} \\
  \cmidrule(lr){3-5}\cmidrule(lr){6-8}
  Method & Additional input
  & ADE $\downarrow$ & FDE $\downarrow$ & PWT $\uparrow$
  & ADE $\downarrow$ & FDE $\downarrow$ & PWT $\uparrow$ \\
  \midrule
  \method{} (ours) & None
  & \bestcell{4.47} & \bestcell{6.23} & \bestcell{76}
  & \bestcell{9.31} & \bestcell{14.88} & \bestcell{54} \\
  MolmoMotion, no language & Null action field
  & 5.66 & 8.90 & 70
  & 9.50 & 16.05 & 53 \\
  Video generation + tracks & None
  & 11.20 & 12.58 & 40
  & 13.82 & 17.65 & 32 \\
  \midrule
  MolmoMotion, with language & Action description
  & 5.93 & 9.38 & 68
  & 9.94 & 17.16 & 51 \\
  Video generation + tracks & Action description
  & 11.99 & 13.57 & 44
  & 13.63 & 16.71 & 29 \\
  \bottomrule
  \end{tabular}%
  }

\end{table}
\subsection{Comparison with baselines}
\label{sec:baselines}

We compare \method{} with future video generation (with Wan-VACE) followed by our 3D track-extraction pipeline (inspired by prior works~\cite{li2025novaflow,kuang2026dex4d,gen2act}) and with concurrent work MolmoMotion~\citep{molmomotion}. We evaluate both baselines without language (i.e. a setting similar to MotionForesight) and with a ground-truth action description (i.e. using more proviledged information than MotionForesight). For SSv2 we use the language annotations provided and on phone videos, the language descriptions are manually written.

\begin{wraptable}{r}{0.60\linewidth}
  \centering
  \scriptsize
  \vspace{-0.8\baselineskip}
  \caption{\footnotesize\textbf{Long-context evaluation on OOD phone videos.}
  The first 50\% of each video is observed and the remaining 50\% is forecast.
  All methods share the same split and evaluation inputs; the language-conditioned MolmoMotion variant and the Video generation model additionally receive the ground-truth action description in language where specified.}
  \label{tab:long_context_ood}
  \begin{tabular*}{\linewidth}{@{\extracolsep{\fill}}lccc@{}}
  \toprule
  & \multicolumn{3}{c}{OOD phone videos} \\
  \cmidrule(lr){2-4}
  Method & ADE $\downarrow$ & FDE $\downarrow$ & PWT@5cm $\uparrow$ \\
  \midrule
  \method{} (ours) & \bestcell{10.20} & \bestcell{13.59} & \bestcell{47.4} \\
  MolmoMotion (no lang.) & 11.90 & 16.27 & 41.3 \\
   Videogen + tracks (no lang.) & 13.33 & 15.12 & 20.9 \\
    MolmoMotion (+ lang.) & 12.57 & 18.77 & 37.1 \\
  Videogen + tracks (+ lang.) & 12.91 & 14.55 & 17.7 \\
  \bottomrule
  \end{tabular*}
  \vspace{-0.5\baselineskip}
\end{wraptable}

Table~\ref{tab:baselines} shows that \method{} performs best on every SSv2 and OOD metric, despite using neither language nor action labels. With the longer observed context in Table~\ref{tab:long_context_ood}, \method{} substantially outperforms MolmoMotion: its ADE is 10.20\,cm, compared with 11.90\,cm without language and 12.57\,cm with language, with corresponding gains in FDE and PWT@5cm. This advantage---despite MolmoMotion's much larger, language-paired training corpus~\citep{molmomotion}---suggests that strong geometric grounding is especially useful for converting richer observed context into metrically consistent future motion. Video generation followed by track extraction performs substantially worse than either geometry-aware method. This supports our central motivation: a pretrained video generator contains useful temporal priors, but visually plausible future pixels do not necessarily preserve point correspondence or metric 3D motion. Among methods that predict geometry explicitly, \method{} benefits from reasoning jointly over observed RGB, reconstructed pointmaps, and a tracking-adapted video representation, preserving scene structure while exploiting the backbone's motion prior.

Note that ADE, FDE, and PWT compare a prediction with one realized future, but interaction forecasting is inherently multimodal, and there could be multiple plausible future trajectories. A predicted forecast can therefore disagree with the recorded ground truth while remaining physically and semantically plausible. We consequently treat these metrics as complementary rather than exhaustive and provide side-by-side qualitative comparisons in Appendix Fig.~\ref{fig:baseline_qualitative} and the website.

\subsection{Scaling with human-interaction videos}
\label{sec:scaling}

We train the same model on nested, action-template-stratified subsets containing 1K, 10K, and 40K SSv2 videos. Architecture, initialization, optimization, and evaluation sets are fixed; only the amount and diversity of fine-tuning data change. Quantitatively, the 40K model is strongest overall: it performs best on all three SSv2 metrics and achieves the highest OOD PWT, while the 10K model has slightly lower OOD ADE and FDE. Qualitatively, however, the 40K model is almost always better, producing more coherent and object-aligned future motion than the smaller-data variants. This mismatch is expected for a multimodal task: displacement to a single recorded future can favor a conservative trajectory even when another prediction is more plausible. Overall, the results indicate that greater data diversity improves the learned motion prior, although that gain is not always reflected monotonically by these metrics. Thus, we perform additional analyses in section~\ref{sec:dynamics_metrics}, and qualitatively visualize the results in the project website \href{https://motionforesight.github.io/}{motionforesight.github.io}.

  \begin{table}[t]
  \centering
  \scriptsize
  \caption{\footnotesize\textbf{Scaling with human-interaction videos.} We evaluate models trained on increasingly large,
  action-stratified subsets of SSv2 while holding all other settings fixed.}
  \label{tab:scaling}
  \begin{tabular*}{\linewidth}{@{\extracolsep{\fill}}lcccccc@{}}
  \toprule
  & \multicolumn{3}{c}{SSv2 validation}
  & \multicolumn{3}{c}{OOD phone videos} \\
  \cmidrule(lr){2-4}\cmidrule(lr){5-7}
  Training videos
  & ADE $\downarrow$ & FDE $\downarrow$ & PWT $\uparrow$
  & ADE $\downarrow$ & FDE $\downarrow$ & PWT $\uparrow$ \\
  \midrule
  1K  & 4.81 & 6.57 & 74 & 9.48 & 14.74 & 53 \\
  10K & 4.72 & 6.38 & 73 & \bestcell{8.97} & \bestcell{14.63} & 52 \\
  40K & \bestcell{4.47} & \bestcell{6.23} & \bestcell{76} & 9.31 & 14.88 & \bestcell{54} \\
  \bottomrule
  \end{tabular*}
  \end{table}

\subsection{Qualitative results}
\label{sec:qualitative}

Figure~\ref{fig:qualitative} shows that \method{} predicts smooth, spatially coherent motion conditioned on the observed interaction. The model captures lifting, translation, rotation, constrained sliding, changes in direction, and local nonrigid motion rather than simply extrapolating instantaneous velocity. It also produces plausible forecasts on independently recorded phone videos, despite differences in objects, environments, viewpoints, and capture conditions. This transfer suggests that adapting the pretrained video-and-tracking representation preserves useful motion priors beyond the SSv2 fine-tuning distribution. As expected, deterministic prediction remains challenging when the observed context admits several plausible futures; comparisons with the baselines are shown in Appendix Fig.~\ref{fig:baseline_qualitative}. Please check the website for more detailed qualitative results in diverse scenarios \href{https://motionforesight.github.io/}{motionforesight.github.io}.

\subsection{Motion-conditional dynamics analysis}
\label{sec:dynamics_metrics}

Standard metrics such as ADE, FDE, and PWT remain useful for comparison with prior work, but they compare a deterministic prediction with only one realized future. When the observed interaction admits several plausible outcomes, they can favor a conservative prediction near the starting position. We therefore add motion-conditional diagnostics that explicitly evaluate whether the predicted dynamics of the motion agree with the realized interaction.

\textit{Trajectory-Vector Overlap (TVO)} compares the predicted and ground-truth displacement vector at every future frame. It gives high credit only when a point moves in the correct direction, by the correct amount, and at the correct time; delayed, missing, or excessive motion is penalized. \textit{Velocity-Vector Overlap (VVO)} applies the same overlap to frame-to-frame velocity. It complements TVO by responding more strongly to local changes such as acceleration, turns, stops, and reversals. \textit{MoveF1} asks whether the model moves the correct object points beyond a small motion threshold. Its threshold-free companion, \textit{MoveIoU}, compares the peak predicted and ground-truth excursion of every point and therefore also reflects motion magnitude. \textit{DQS} combines trajectory fidelity and motion placement through the geometric mean of TVO and MoveF1. Finally, the motion ratio $\bar r$ diagnoses magnitude calibration: values below one indicate motion that is too timid, while values above one indicate overshooting. Full mathematical definitions are provided in Appendix~\ref{app:dynamics_metrics}.

\begin{table}[t]
\centering
\scriptsize
\setlength{\tabcolsep}{3pt}
\caption{\footnotesize\textbf{Motion-conditional evaluation on SSv2 unseen clips.} We use $\tau=2$\,cm and a three-frame smoothing window. Darker and lighter blue denote the best and second-best learned result for each metric. TVO/VVO/DQS are evaluated on the 108 clips with at least five ground-truth-moving points.}
\label{tab:dynamics_val}
\begin{tabular*}{\linewidth}{@{\extracolsep{\fill}}lccccccc@{}}
\toprule
Method & TVO $\uparrow$ & VVO $\uparrow$ & MoveF1 $\uparrow$ & MoveIoU $\uparrow$ & DQS $\uparrow$ & $\bar r\!\to\!1$ \\
\midrule
\method{} (40K) & \bestcell{0.231} & \bestcell{0.175} & \bestcell{0.618} & \bestcell{0.448} & \bestcell{0.326} & 0.72  \\
\method{} (10K) & \secondcell{0.167} & 0.127 & 0.582 & 0.388 & 0.237 & 0.57  \\
\method{} (1K) & 0.150 & 0.112 & 0.487 & 0.370 & 0.186 & 0.50 \\
\midrule
MolmoMotion (no language) & 0.101 & 0.078 & 0.585 & 0.258 & 0.195 & \secondcell{1.27} \\
MolmoMotion (language) & 0.122 & 0.089 & 0.586 & 0.269 & 0.225 & 1.46 \\
\midrule
Video generation + tracks (no language) & 0.165 & \secondcell{0.138} & 0.568 & \secondcell{0.435} & 0.228 & 0.52 \\
Video generation + tracks (language) & 0.157 & 0.137 & \secondcell{0.613} & 0.415 & \secondcell{0.256} & \bestcell{0.90} \\
\bottomrule
\end{tabular*}
\end{table}

Table~\ref{tab:dynamics_val} shows that \method{} (40K) leads all five motion-quality metrics, and its TVO and DQS improve monotonically from 1K to 40K training videos. Video generation followed by tracking obtains relatively strong motion placement but lower TVO and DQS, indicating that it can identify which points should move without preserving their metric trajectories as accurately. The 40K model still under-predicts total motion ($\bar r=0.72$), leaving magnitude calibration as an important direction for improvement. Qualitative results of predictions shown in the project website further support these quantitative analyses \href{https://motionforesight.github.io/}{motionforesight.github.io}.

\section{Discussion, Limitations, and Conclusion}

In this work, we studied future 3D scene-flow prediction from short monocular videos of human--object interaction. We showed that a tracking-adapted video model can be converted from retrospective reconstruction into prospective prediction by masking future observations and training only a lightweight adapter, while preserving the pretrained video and tracking components. The resulting reference-anchored 3D tracks expose the metric variable needed for physical reasoning without restricting the object to a single rigid 6-DoF pose, allowing one representation to describe translation, rotation, articulated motion, and local deformation. Our quantitative, motion-conditional, and qualitative results show that the prediction model captures meaningful interaction dynamics, scales with additional human video, and transfers to phone captures on home and office scenes.

Several challenges remain. The current model is deterministic and predicts only one future, although a short interaction prefix may support multiple physically plausible outcomes; both conventional displacement metrics and our motion-conditional diagnostics still compare that prediction with a single recorded trajectory. Supervision is also pseudo ground truth, so errors from monocular depth, camera estimation, segmentation, and tracking can propagate into both training and evaluation, while inference remains sensitive to errors in the estimated observed pointmaps. Finally, training is limited to 40K SSv2 manipulation clips. Although the OOD phone experiments demonstrate transfer across objects, environments, and viewpoints, they do not establish robustness to substantially different regimes such as head-mounted egocentric video, very large camera motion, longer-horizon interactions, or non-manipulation dynamics.

These limitations suggest clear directions for future work. Multi-hypothesis or probabilistic forecasting could represent alternative outcomes and expose calibrated uncertainty, paired with evaluation protocols that measure both plausibility and coverage across multiple valid futures. More accurate or jointly learned geometry, segmentation, and tracking could reduce dependence on a fixed pseudo-label pipeline, while broader training data could cover egocentric viewpoints, stronger camera motion, longer interactions, and more diverse physical processes. A further step is to connect the predicted 3D tracks to downstream planning and control, testing whether an object-centered motion prior learned from passive human video improves embodied decision making. Overall, \method{} provides evidence that large video priors can be efficiently redirected from rendering future appearance toward forecasting explicit, actionable 3D dynamics without language or action supervision.

\bibliographystyle{unsrtnat}
\bibliography{references}

\clearpage
\appendix

\section{Implementation Details}
\label{app:implementation_details}

Here we discuss implementation details about the method, and additional details about the experiment setup. We also showcase additional results to help understand our method and the baselines.

\subsection{Default model configuration}

\begin{table}[H]
\centering
\scriptsize
\caption{\footnotesize\textbf{Default model configuration.} The method starts from a pretrained video DiT, uses a learned point-track latent interface as the geometric output representation, and trains only a small future adapter.  Language is held at a null context and is not an input.}
\label{tab:model_config}
\begin{tabularx}{0.95\linewidth}{lX}
\toprule
Component & Setting \\
\midrule
Input horizon & \DefaultK{} observed frames, \DefaultN{} predicted future frames, \DefaultT{} total frames \\
Input signals & RGB context frames plus estimated observed pointmaps; no language, action, robot state, or future-frame input \\
Reference points & Object query points sampled from the frame-0 object mask \\
Coordinate frame & Last-observed camera frame for the default camera-subtracted model; reference object points still originate from frame 0 \\
Video backbone & \wan{} T2V DiT with temporal RoPE; used as a single-step latent regressor \\
Track interface & Tracking-adapted point-track latent interface, instantiated with \trackcraft{} encoders/decoders and frozen rank-1024 tracking LoRA \\
Frozen components & Base video DiT, released tracking LoRA, RGB VAE, pointmap VAE, track decoder, visibility decoder, and null text context \\
Trainable adapter & Fresh rank-32 LoRA on self-attention projections \((q,k,v,o)\) \\
Other trainable parameters & Future RGB mask latent, future pointmap mask latent, patch embedding, and output head \\
Trainable parameter count & \TrainableParams{} \\
Resolution & \ModelRes{} \\
Training loss & Decoded coordinate-space MSE at query points, scaled by 10 in implementation; future weight 1.0, observed weight 0.25 \\
\bottomrule
\end{tabularx}
\end{table}

\subsection{Data and pseudo-label generation}

The main training corpus is built from human-object interaction videos.  For each raw video, preprocessing selects a \DefaultT-frame clip centered around an interaction.  The pipeline searches for visible hand-object interaction, advances to an anchor frame near contact, segments the manipulated object, and writes a fixed-length clip.  Monocular depth and camera estimates are then computed for the clip, producing pointmaps and camera transforms.  Finally, the dense 3D tracking stack is run on the complete clip to produce pseudo-ground-truth future tracks.

The distinction between label construction and predictor input is important.  The full clip is used offline to obtain supervision, but the forecasting model never receives future frames as input.  During training and inference, the context slots for \(t<\Kctx\) contain observed RGB and pointmap latents, while the future slots for \(t\geq\Kctx\) contain only learned mask latents.

The loader presents dense and sparse supervision through a common query-point interface.  Dense tracking outputs are converted into sampled object query points from the reference mask.  Sparse tracking outputs use their native valid query points.  In both cases, the loss is evaluated only at query points with valid pseudo-ground-truth supervision.

\subsection{Coordinate frame and normalization}

The default model expresses pointmaps and track targets in the camera coordinate frame of the last observed image, \(t=T_1-1\).  Let \(\bar{P}_t\) denote a pointmap in its original camera coordinate frame, and let \(G_{t\rightarrow T_1-1}\) denote the estimated transform into the last-observed camera frame.  The model uses
\begin{equation}
    P_t = G_{t\rightarrow T_1-1}\bar{P}_t .
\end{equation}
For observed frames, this transformation is applied before pointmap encoding.  For future frames, the transformed tracks are used only as training targets.  At inference time, no future geometry is available.

This coordinate choice removes a large part of camera motion from the prediction problem.  The model still predicts the future 3D position of reference-frame object points, but the coordinates are expressed relative to the last observed camera rather than the original frame-0 camera.  Earlier dense-only variants used frame-0 coordinates; the last-observed camera frame is the default because it makes the target more object-motion-centered.

Pointmap normalization statistics are computed from observed pointmaps only.  This keeps training consistent with inference, where the model has access to no future pointmaps.  The same observed-prefix statistics are used to normalize the observed pointmap latents and to define the scale of the decoded coordinate loss.

\subsection{Training objective details}

The default training objective is the decoded coordinate-space loss in Eq.~\ref{eq:decoded_loss}.  We use
\[
    \lambda_{\mathrm{obs}} = 0.25,
    \qquad
    \lambda_{\mathrm{fut}} = 1.0 .
\]
The coordinate MSE is multiplied by 10 in the implementation.  The future term is the primary objective, while the observed term keeps the adapted model aligned with the inherited tracking interface.

The track decoder is frozen, but gradients are allowed to pass through it into the trainable future adapter, patch embedding, output head, and mask latents.  To fit the decoded loss at \ModelRes{}, the implementation decodes one frame at a time and uses gradient checkpointing.  This makes the decoded metric loss feasible without updating the large video or tracking components.

\subsection{Forecasting pass}

At inference time, the model receives only the observed prefix.  The forward pass is:

\begin{enumerate}
    \item Estimate observed pointmaps \(P_{0:\Kctx-1}\) from the RGB context frames.
    \item Transform the observed pointmaps into the last-observed camera frame.
    \item Encode observed RGB frames and pointmaps into visual-geometry latents.
    \item Replace all future RGB and pointmap context latents with learned mask latents.
    \item Repeat the reference-frame query latent across all \(T\) time steps.
    \item Run the frozen video DiT with the frozen tracking adapter and the trained future adapter.
    \item Decode residual-track latents with the frozen track decoder.
    \item Add the decoded residuals to the reference point positions to obtain future 3D tracks.
\end{enumerate}

\subsection{Evaluation metrics}

For held-out validation clips with pseudo-ground-truth tracks, we evaluate future average displacement error and final displacement error over valid object query points.  Let \(\mathcal{T}_{\mathrm{fut}}=\{\Kctx,\ldots,T-1\}\).  The future ADE is
\begin{equation}
    \mathrm{ADE}
    =
    \frac{
    \sum_{t\in\mathcal{T}_{\mathrm{fut}}}
    \sum_{q}
    v_{tq}
    \left\|
    \hat{X}_t(q)-X_t(q)
    \right\|_2
    }{
    \sum_{t\in\mathcal{T}_{\mathrm{fut}}}
    \sum_q
    v_{tq}
    + \epsilon
    } .
\end{equation}
The future FDE is computed at the final predicted frame:
\begin{equation}
    \mathrm{FDE}
    =
    \frac{
    \sum_q
    v_{T-1,q}
    \left\|
    \hat{X}_{T-1}(q)-X_{T-1}(q)
    \right\|_2
    }{
    \sum_q
    v_{T-1,q}
    + \epsilon
    } .
\end{equation}
Both metrics are reported in metric 3D space. For the OOD phone videos, we evaluate against pseudo-ground-truth tracks extracted from each complete clip and supplement the metrics with qualitative comparisons. Because each clip records only one of several plausible futures, ADE and FDE should not be interpreted as complete measures of forecast quality: a prediction can differ from the recorded trajectory while remaining physically plausible and consistent with the observed interaction.

\section{Motion-Conditional Dynamics Metrics}
\label{app:dynamics_metrics}

The following diagnostics measure predicted motion relative to the last observed frame rather than absolute position. Let $O=T_1$ be the number of observed frames, and let $p_{t,i},\hat p_{t,i}\in\mathbb{R}^3$ denote the ground-truth and predicted position of point $i$. We define independently anchored displacements
\begin{equation}
\Delta_{t,i}=p_{t,i}-p_{O-1,i},
\qquad
\hat\Delta_{t,i}=\hat p_{t,i}-\hat p_{O-1,i},
\qquad t\geq O .
\end{equation}
This removes constant offsets at the forecast boundary and isolates the subsequent dynamics. To reduce one-frame noise in the monocular pseudo-ground-truth tracks, both series are smoothed with a centered three-frame moving average: $g_{t,i}=\operatorname{sm}(\Delta)_{t,i}$ and $h_{t,i}=\operatorname{sm}(\hat\Delta)_{t,i}$.

\paragraph{Moving points.}
A point is considered moving if its peak excursion from the anchor exceeds $\tau$:
\begin{equation}
e_i=\max_{t\geq O}\|g_{t,i}\|_2,
\qquad
\hat e_i=\max_{t\geq O}\|h_{t,i}\|_2,
\qquad
\mathcal M=\{i:e_i\geq\tau\},
\quad
\hat{\mathcal M}=\{i:\hat e_i\geq\tau\}.
\end{equation}
Peak excursion counts motion that later returns to its starting position. We use $\tau=2$\,cm, which remains above the smoothed pseudo-track jitter floor while retaining 108 of 150 valid SSv2 clips for fidelity evaluation; results are also checked at $\tau\in\{1,5,10\}$\,cm. A clip is fidelity-eligible when $|\mathcal M|\geq5$.

\paragraph{Trajectory-Vector Overlap.}
For each ground-truth-moving point, TVO measures frame-aligned agreement in direction and magnitude:
\begin{equation}
\mathrm{TVO}_i=
\frac{
\sum_{t\geq O}
\left[\cos(h_{t,i},g_{t,i})\right]_+
\min\!\left(\|h_{t,i}\|_2,\|g_{t,i}\|_2\right)
}{
\sum_{t\geq O}
\max\!\left(\|h_{t,i}\|_2,\|g_{t,i}\|_2\right)+\epsilon
},
\end{equation}
where $[x]_+=\max(x,0)$ and the cosine term is defined as zero if either vector has zero norm. Missing, delayed, excessive, or directionally incorrect motion leaves unmatched magnitude in the denominator. We average over moving points within each eligible clip and then average clips equally.

\paragraph{Velocity-Vector Overlap.}
TVO compares displacement chords from the anchor, which can change slowly along a curved path. VVO applies the same overlap to smoothed temporal velocities,
\begin{equation}
v_{t,i}=g_{t,i}-g_{t-1,i},
\qquad
\hat v_{t,i}=h_{t,i}-h_{t-1,i},
\qquad g_{O-1,i}=h_{O-1,i}=0,
\end{equation}
and is aggregated using the TVO equation with $(h,g)$ replaced by $(\hat v,v)$. It therefore emphasizes whether a prediction turns, reverses, accelerates, or stops at the correct time.

\paragraph{Motion placement and magnitude.}
MoveF1 is the point-set F1 score between $\hat{\mathcal M}$ and $\mathcal M$. If neither set contains a moving point, the score is one; if exactly one set is empty, it is zero. Its threshold-free companion compares peak excursions for all points:
\begin{equation}
\mathrm{MoveIoU}=
\frac{\sum_i\min(\hat e_i,e_i)}
{\sum_i\max(\hat e_i,e_i)+\epsilon}.
\end{equation}
We additionally report the clip-averaged motion ratio
\begin{equation}
r_c=
\frac{\sum_{i,t\geq O}\|h_{t,i}\|_2}
{\sum_{i,t\geq O}\|g_{t,i}\|_2+\epsilon},
\qquad
\bar r=\frac{1}{|\mathcal C|}\sum_{c\in\mathcal C}r_c,
\end{equation}
where $\bar r=1$ indicates calibrated total motion, values below one indicate under-prediction, and values above one indicate over-prediction. For a compact summary, we compute $\mathrm{DQS}_c=\sqrt{\mathrm{TVO}_c\,\mathrm{MoveF1}_c}$ on each fidelity-eligible clip before averaging. We treat TVO and MoveF1 as the primary components because the composite can hide whether an error comes from trajectory fidelity or motion placement.

These diagnostics still compare against one realized future and therefore complement the qualitative evaluations of plausible alternative outcomes.

\begin{figure}
    \centering
    \includegraphics[width=\linewidth]{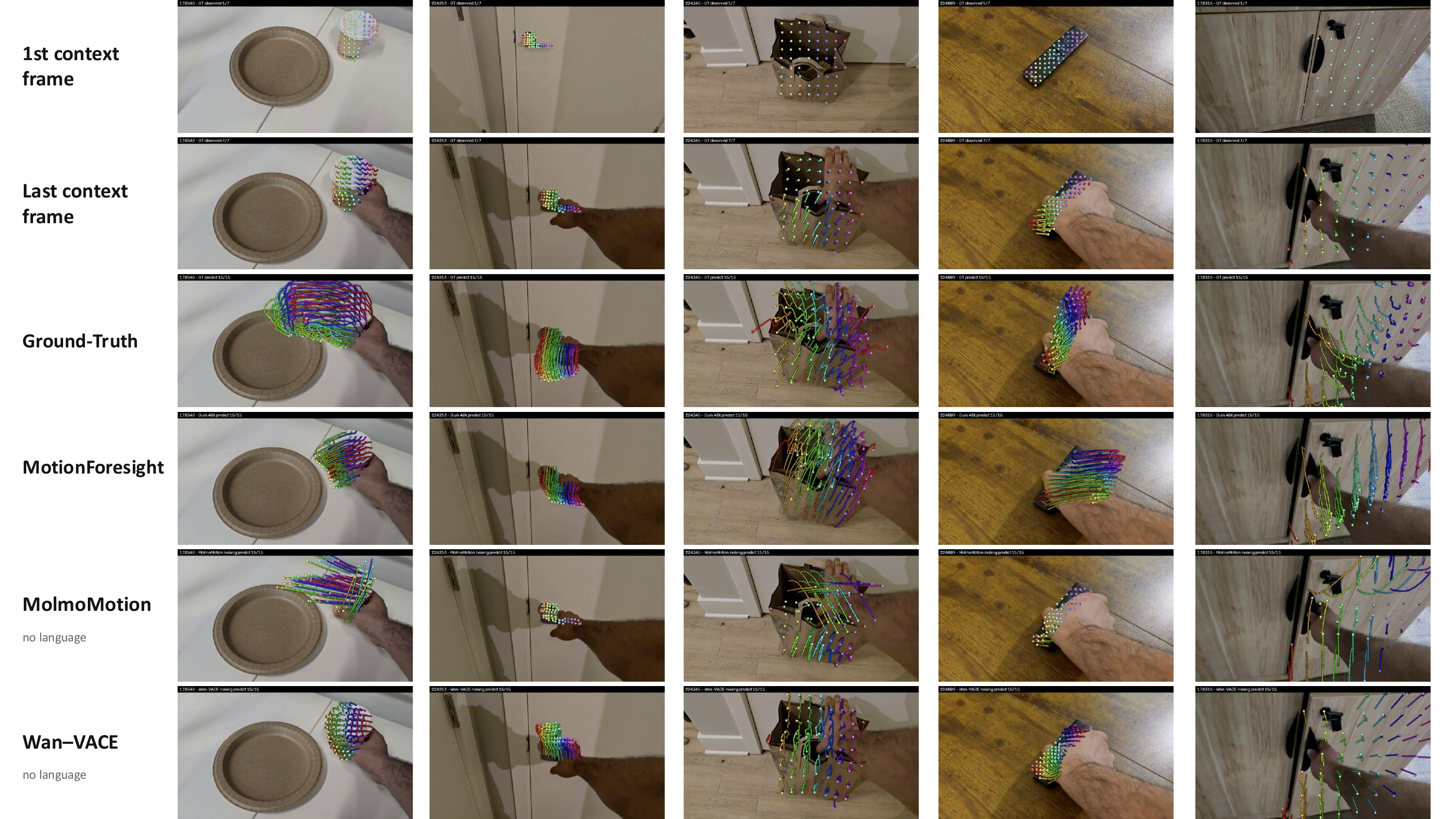}
    \caption{\textbf{Qualitative comparison with baselines.} We show the observed context, the recorded future tracks, and predictions from \method{}, MolmoMotion, and video generation followed by tracking. These examples complement the single-ground-truth trajectory metrics by revealing motion coherence and plausible alternative futures.}
    \label{fig:baseline_qualitative}
\end{figure}

\end{document}